\documentclass[letterpaper]{article} 
\usepackage{aaai2027}  
\usepackage[hyphens]{url}  
\usepackage{graphicx} 
\usepackage{natbib}  
\usepackage{caption} 
\usepackage{amsmath}
\usepackage{amsfonts}
\usepackage{adjustbox}
\usepackage{booktabs}
\usepackage{multirow}
\usepackage[table]{xcolor}
\usepackage{tabularx}
\usepackage{array}
\newcolumntype{Y}{>{\centering\arraybackslash}X}
\newcolumntype{L}[1]{>{\raggedright\arraybackslash}p{#1}}
\usepackage{algorithm}
\usepackage{algorithmic}

\usepackage{newfloat}
\usepackage{listings}
\DeclareCaptionStyle{ruled}{labelfont=normalfont,labelsep=colon,strut=off} 
\floatstyle{ruled}
\newfloat{listing}{tb}{lst}{}
\floatname{listing}{Listing}

\usepackage{booktabs}

\title{Sparse Evidence Can Suffice: Agentic Evidence Seeking for Multimodal Video Misinformation Detection}
\author{
    Haochen Zhao\textsuperscript{\rm 1,2},
    Yongxiu Xu\textsuperscript{\rm 1,2}\corresponding,
    Xinkui Lin\textsuperscript{\rm 1,2},
    Dong Xie\textsuperscript{\rm 3},\\
    Jiarui Lu\textsuperscript{\rm 1,2},
    Yuqi Qian\textsuperscript{\rm 1,2},
    Yubin Wang\textsuperscript{\rm 1,2},
    Hongbo Xu\textsuperscript{\rm 1},
    Gaopeng Gou\textsuperscript{\rm 1}
}

\affiliations{
    \textsuperscript{\rm 1}
    Institute of Information Engineering, Chinese Academy of Sciences,
    Beijing, China\\
    \textsuperscript{\rm 2}
    School of Cyber Security, University of Chinese Academy of Sciences,
    Beijing, China\\
    \textsuperscript{\rm 3}
    Baidu,
    Beijing, China\\
    zhaohaochen@iie.ac.cn
}

\begin{document}

\maketitle
\begin{abstract}
Multimodal video misinformation detection is commonly formulated as a holistic video-understanding task, where the entire video and its associated content are processed and judged in a single pass. However, real-world misinformation often exhibits a sparse and compositional evidence structure: a reliable decision may depend on only a few coupled clues, while most video content contributes limited additional information. Exhaustive multimodal reasoning may therefore introduce substantial redundancy and obscure decisive evidence. This motivates decoupling evidence acquisition from verification: first identifying sparse, decision-relevant clues and then judging veracity based on the acquired evidence. Accordingly, we propose \textbf{SIEVE}, a framework for \textbf{S}parse \textbf{I}nteractive \textbf{E}vidence \textbf{V}erification via \textbf{E}xtraction in multimodal video misinformation detection. An evidence-seeking agent actively explores the available multimodal evidence and constructs a compact evidence package, which is then used by a verifier to determine veracity. The agent is trained with supervised evidence-seeking trajectories and an evidence-aware reinforcement learning objective that promotes informative evidence acquisition while discouraging unnecessary or invalid interactions. Experiments on multiple video misinformation benchmarks show that SIEVE consistently outperforms the evaluated baselines and supports reliable verification using compact evidence packages. Moreover, the resulting acquisition process provides an explicit and inspectable evidence trail, improving the transparency and groundedness of multimodal misinformation detection.
\end{abstract}


\section{Introduction}
\begin{figure}[t]
    \centering
    \includegraphics[width=\columnwidth]{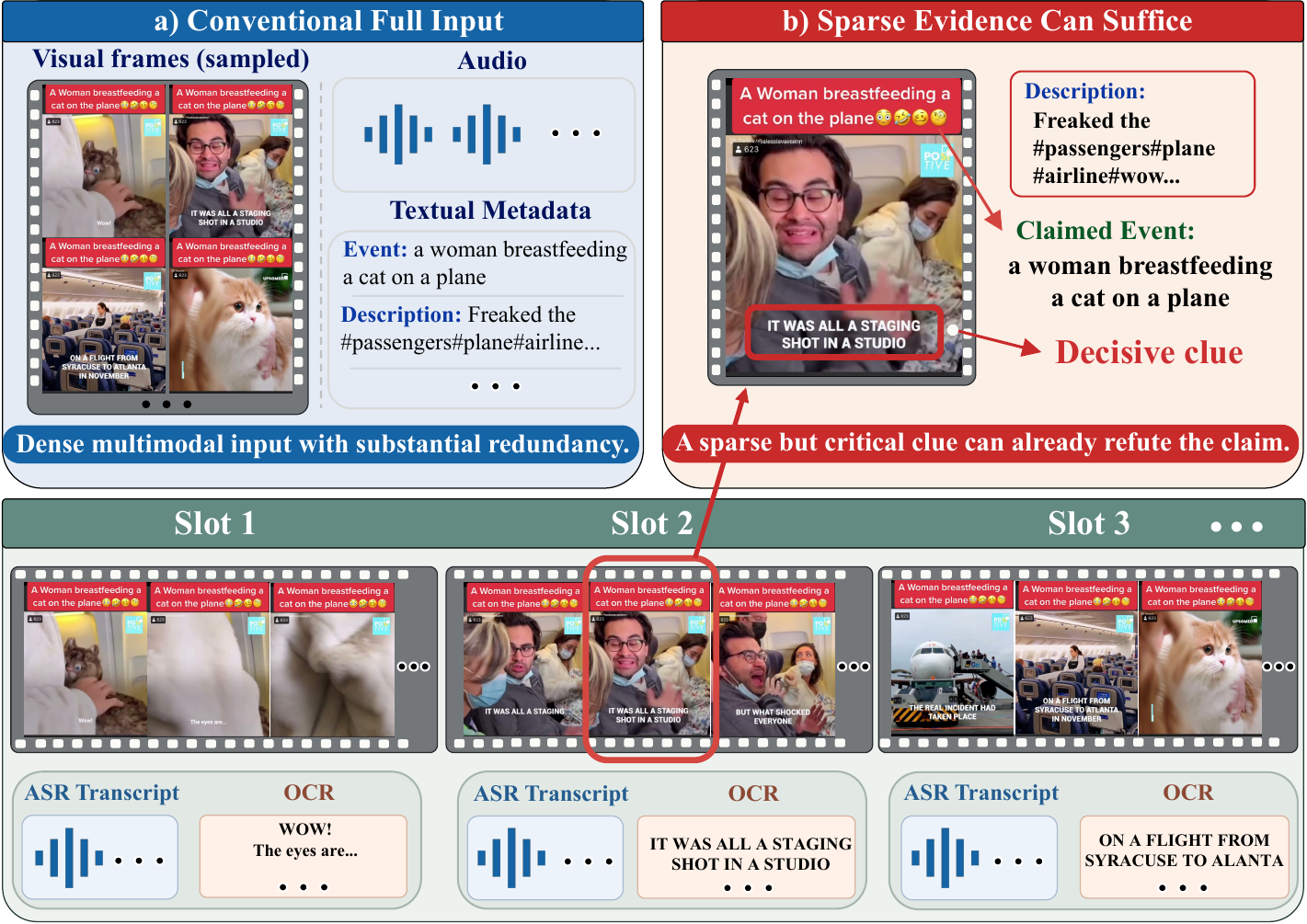}
    \caption{An example illustrating that sparse, claim-relevant evidence can suffice for video claim verification.}
    \label{fig:motivation}
\end{figure}

Videos have become a dominant medium for online information consumption, especially on short-video platforms such as TikTok and YouTube, where user-generated videos can be rapidly created and widely disseminated~\cite{bu2023combating,venkatagiri2023challenges}. However, the same ease of video creation and dissemination also accelerates the spread of misinformation, threatening the credibility of online information ecosystems and public discourse~\cite{bu2023combating,lazer2018science}. Since the volume and velocity of online videos far exceed the capacity of manual fact-checking, automatic video misinformation detection has become an increasingly important research problem~\cite{venkatagiri2023challenges,bu2023combating}.

Compared with text- or image-based misinformation, video misinformation is more challenging because videos contain heterogeneous visual, acoustic, textual, and social signals, as well as temporal dynamics and subtle cross-modal inconsistencies~\cite{bu2023combating,shang2021multimodal,liu2023covidvts}. Early video misinformation detection methods mainly rely on task-specific encoders and multimodal fusion strategies, such as modeling cross-modal correlations, social context, creative-process patterns, or neighborhood relations among event-related videos~\cite{choi2021topic,liu2023covidvts,qi2023fakesv,li2025learning,bu2024fakingrecipe,qi2023twoheads}. Although these methods have achieved promising results, they are typically optimized as end-to-end discriminative classifiers, offering limited support for explicit, interpretable, and flexible multi-step evidence reasoning in open-world verification scenarios~\cite{wang2025fakesvvlm,zheng2025predictions,wang2026csi,lang2026nip}.

Recently, vision-language models and multimodal large language models have shown strong potential for fake short-video news detection by leveraging their broad multimodal knowledge and reasoning capabilities~\cite{zheng2025predictions,wang2025fakesvvlm,hong2025following,zhang2025factr1}. Some methods introduce chain-of-thought reasoning to improve explainability~\cite{hong2025following}, while others further explore long reasoning trajectories and reinforcement learning for video misinformation detection~\cite{zhang2025factr1,li2026factguard}. Nevertheless, most existing methods still follow a single-pass or fixed-depth inference paradigm, where the model directly predicts a label from pre-extracted inputs. When critical evidence is sparse, fragmented, or insufficient, such models may rely on internal assumptions rather than actively acquiring targeted evidence, leading to unreliable or poorly grounded decisions~\cite{yao2023react,zhang2023relevance,li2026factguard,wang2026csi}.

As shown in Figure~\ref{fig:motivation}, a video claim can sometimes be resolved from sparse and localized evidence without encoding the entire evidence space. The datasets provide Event and Description metadata, but SIEVE uses only the Description field as the initial claim context and excludes the Event field from all model inputs. In the example, one OCR clue, ``A WOMAN BREASTFEEDING A CAT ON THE PLANE,'' identifies the event asserted by the video post, while another, ``IT WAS ALL A STAGING SHOT IN A STUDIO,'' reveals that the depicted scene was staged. The former establishes what is being claimed, whereas the latter provides the evidence needed to refute it; considered together, these complementary clues are sufficient to reject the claim. This example highlights a challenge beyond multimodal fusion: how to identify and combine the sparse, decisive clues needed to resolve a claim within a largely redundant evidence space, where exhaustive evidence consumption is unnecessary and may obscure the evidence that determines the judgment.

Accordingly, we formulate video misinformation detection as a budget-constrained evidence-seeking and verification process, where an acquisition budget encourages the agent to prioritize small sets of potentially decisive and complementary clues rather than exhaustively reading all available content. Specifically, the platform-provided Description field is treated as the initial claim context, while the dataset-provided Event field is not used. An evidence-seeking agent selectively reads video-derived evidence channels, including visual frames, OCR, and ASR, and constructs a compact evidence package in which complementary clues can be jointly interpreted. A verifier then makes the judgment from the Description and the acquired evidence. This design decouples evidence acquisition from verification and exposes both the acquired evidence and its compositional basis for inspection.

Our main contributions are summarized as follows:

\begin{itemize}

\item We formulate multimodal video misinformation detection as a \textbf{selective evidence acquisition and verification problem} and empirically investigate whether a compact subset of video-derived evidence can support reliable verification within a bounded multimodal context.

\item We propose \textbf{SIEVE}, an agentic framework that actively seeks potentially decisive and complementary multimodal evidence and constructs a \textbf{compact evidence package} for verification. The evidence agent is trained with supervised trajectories and evidence-aware reinforcement learning for grounded evidence acquisition.

\item We conduct experiments on \textbf{multiple video misinformation benchmarks}. SIEVE consistently outperforms the evaluated baselines and reaches near-saturated Macro-F1 with a moderate evidence budget. Its decoupled design also provides an explicit and inspectable evidence acquisition trail.

\end{itemize}

\section{Related Work}

Existing video misinformation detection methods usually formulate the task as holistic multimodal classification, jointly encoding visual, textual, acoustic, and social signals for prediction. Early studies investigate topic modeling, adversarial learning, and domain-specific multimodal fusion for fake news or COVID-19 misinformation videos~\cite{choi2021topic,shang2021multimodal,liu2023covidvts}. Later works build short-video benchmarks and introduce richer contextual modeling. FakeSV incorporates multimodal content and social context for fake short-video news detection~\cite{qi2023fakesv}, while subsequent methods exploit neighborhood relations among event-related videos~\cite{qi2023twoheads}, creative-process cues~\cite{bu2024fakingrecipe}, or heterogeneous social latent structures~\cite{li2025learning}. Despite these advances, most methods process pre-collected video inputs through a single discriminative pipeline, offering limited support for explicitly identifying which sparse multimodal clues are sufficient for grounded verification.

Recent works leverage vision-language models and multimodal large language models to improve semantic reasoning and explainability in misinformation detection. Rationale-augmented VLM methods move from direct prediction to analysis-oriented detection~\cite{zheng2025predictions}, while FakeSV-VLM adapts VLMs through progressive experts and event-level consistency checking~\cite{wang2025fakesvvlm}. ExMRD introduces a refining--retrieving--reasoning chain to produce explainable micro-video rumor predictions~\cite{hong2025following}, and Fact-R1 further combines long-chain reasoning, preference alignment, and reinforcement learning for video misinformation detection~\cite{zhang2025factr1}. More recent agentic frameworks, such as CSI and FactGuard, use multi-agent investigation or tool-augmented reasoning to strengthen evidence acquisition and decision reliability~\cite{wang2026csi,li2026factguard}. Retrieval-guided adaptation has also been explored for test-time fake news video detection~\cite{lang2026nip}. Despite these advances, most methods still reason over preconstructed global inputs or externally retrieved information, rather than learning how to seek compact, localized, and claim-relevant evidence from dense multimodal video content.

Our work is also related to evidence-centered fact verification and selective perception. Textual fact verification has long emphasized the importance of retrieving supporting evidence before making a claim-level judgment~\cite{thorne2018fever}, and later work shows that evidence retrieval should be optimized for downstream verification utility rather than relevance alone~\cite{zhang2023relevance}. In language-agent research, ReAct demonstrates that interleaving reasoning and actions enables models to acquire missing information instead of relying on internal assumptions~\cite{yao2023react}. In video understanding, adaptive frame selection methods show that many video tasks can be solved by observing only a subset of informative frames~\cite{wu2019adaframe}. Unlike these lines of work, SIEVE studies video misinformation detection as a budgeted interactive evidence-seeking problem. It uses only the platform-provided Description field as the initial claim context, explicitly excludes the dataset-provided Event field, and actively acquires compact, claim-relevant evidence from OCR, ASR, and visual channels before evidence-based verification, shifting the task from fixed-input classification to budgeted active evidence acquisition.

\begin{figure*}[t]
    \centering
    \includegraphics[width=\textwidth]{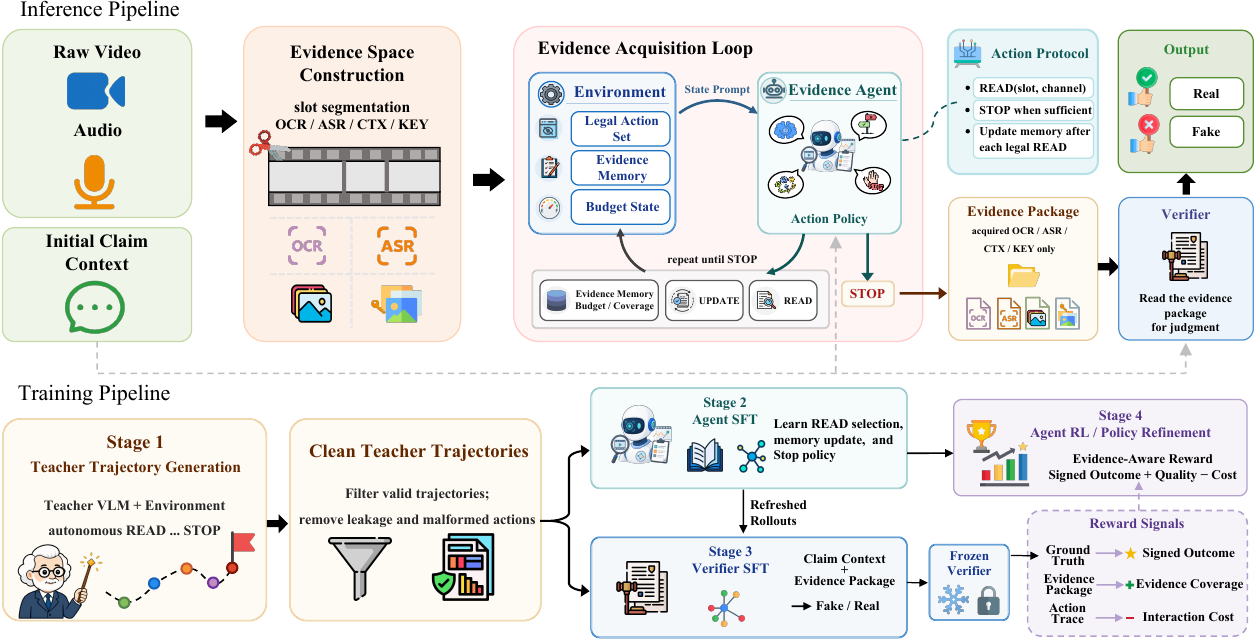}
    \caption{Overall framework of the proposed method.}
    \label{fig:framework}
\end{figure*}

\section{Method}

Additional details on evidence-space construction, training, reward design, and policy optimization are provided in the supplementary material.

\subsection{Overview}
Figure~\ref{fig:framework} presents SIEVE. Given a video and its Description, SIEVE constructs a structured multimodal evidence space. The Evidence Agent receives a one-time slot overview of thumbnails and channel availability for navigation only, then selectively reads evidence under a budget and emits $\mathrm{STOP}$ once sufficient. The overview is excluded from the evidence package and Verifier input. The Verifier predicts from the Description and acquired evidence, enabling compact, evidence-grounded verification.

\subsection{Problem Formulation}

Each instance is $x=(\mathcal{V},\mathcal{A},c)$, where $\mathcal{V}$ and $\mathcal{A}$ are the visual and audio streams and $c$ is the fixed textual context. Both datasets provide Event and Description fields, while FakeSV additionally provides comments. SIEVE uses only Description, i.e., $c=c_{\mathrm{description}}$, excluding Event and Comment from all model inputs. For cross-dataset evaluation on FakeVV, the associated title is used, i.e., $c=c_{\mathrm{title}}$. The task is to predict $y\in\mathcal{Y}=\{\mathrm{Fake},\mathrm{Real}\}$. Conventional methods typically predict directly from the complete multimodal input. In contrast, we formulate detection as active evidence acquisition followed by evidence-based verification. The visual stream is divided into $K$ temporal slots $\mathcal{S}(\mathcal{V})=\{s_i\}_{i=1}^{K}$, each containing available channels from $\mathcal{M}=\{\mathrm{OCR},\mathrm{ASR},\mathrm{CTX},\mathrm{KEY}\}$. The evidence space is
\begin{equation}
    \mathcal{E}(\mathcal{V},\mathcal{A})
    =
    \{e_{i,m}\mid i\in\{1,\ldots,K\},\ m\in\mathcal{M}_i\},
\end{equation}
where $\mathcal{M}_i\subseteq\mathcal{M}$ denotes the channels available in slot $s_i$. Rather than accessing all detailed contents of $\mathcal{E}$, SIEVE sequentially constructs an ordered evidence package $\mathcal{P}_T$ and predicts
\begin{equation}
    \hat{y}=\operatorname{parse}\!\left(g_{\phi}(c,\mathcal{P}_T)\right),
\end{equation}
where $g_{\phi}$ is the autoregressive Verifier.

\subsection{Evidence Space Construction}
\label{sec:evidence-space}

SIEVE builds a slot-based evidence space from each video's audiovisual content. TransNetV2~\cite{soucek2020transnetv2} detects shot boundaries; short shots are merged and long ones subdivided. If over 10 slots remain, adjacent low-motion segments are merged. Each slot may contain OCR, ASR, CTX, and KEY evidence. PaddleOCR~\cite{du2020ppocr} extracts OCR text ranked by temporal, spatial, and recognition-quality cues, while Whisper~\cite{radford2023whisper} produces slot-aligned transcripts. DINOv2 ViT-S/14~\cite{oquab2024dinov2} encodes visual candidates: CTX contains representative frames, whereas KEY contains one or two complementary frames selected by visual distinctiveness and OCR relevance. Before the first action, the environment provides a one-time navigation overview with one thumbnail and channel-availability indicators per slot; it is excluded from the evidence package and Verifier input. Channel state is tracked by $\mathbf{C}_t(i,m)\in\{\mathtt{-},\mathtt{x},\mathtt{v}\}$ for unavailable, unread, and read. Legal READs are $\mathtt{x}$ entries allowed by the remaining budget. Visual READs are capped at 24 accumulated CTX/KEY images, excluding actions that exceed the cap. A legal READ reveals the evidence and changes its state to $\mathtt{v}$. Repeated, unavailable, or malformed actions immediately terminate the rollout, add no evidence, and incur an RL penalty.

\subsection{Interactive Evidence Acquisition}

We formulate evidence acquisition as a budgeted sequential decision process. At step $t$, the agent observes $o_t=(c,\omega,h_t,\mu_t,\mathbf{C}_t,b_t)$, where $c$ is the claim context, $\omega$ is the one-time overview provided at $t=1$, $h_t$ stores acquired evidence and source identifiers, $\mu_t$ is evidence memory, $\mathbf{C}_t$ tracks channel state, and $b_t$ records the remaining READ budget, accumulated visual-image count, and visual READs blocked by the 24-image cap. The agent generates $z_t\sim\pi_\theta(\cdot\mid o_t)$ containing an action and, when applicable, a concise non-verdict memory update for the preceding legal READ. The environment stores the memory update, parses the command, and validates it against the legal read-action set $\mathcal{A}^{\mathrm{read}}_t$. A legal $\mathrm{READ}(\mathrm{slot}_{\mathrm{sid}},\{\mathrm{channel}\})$ reveals $e_{\mathrm{sid},\mathrm{channel}}$ and updates the package, history, coverage, and budget; the new evidence is summarized in the next response. $\mathrm{STOP}$ terminates acquisition. Repeated, unavailable, or malformed commands are invalid attempts: they add no evidence, immediately terminate the current acquisition rollout, and incur a behavioral penalty during reinforcement learning. Acquisition ends upon $\mathrm{STOP}$, budget exhaustion, or invalid action. The terminal package $\mathcal{P}_T$ contains only legally acquired evidence and source identifiers, excluding memory and reasoning traces.

\subsection{Evidence-Based Verification}

Given $c$ and $\mathcal{P}_T$, the Verifier autoregressively generates label tokens:
\begin{equation}
\tilde{\mathbf{y}}=g_{\phi}(c,\mathcal{P}_T), \qquad
\hat{y}=\operatorname{parse}(\tilde{\mathbf{y}}).
\end{equation}
The Verifier receives only the claim context and legally acquired evidence, without access to unread evidence or interaction states such as coverage, budget, or agent memory. This restricted interface encourages evidence-grounded prediction rather than reliance on procedural artifacts. The Verifier is a smaller VLM trained with LoRA and frozen during policy refinement.

\subsection{Training}

SIEVE is trained in four stages: teacher trajectory generation, agent supervised fine-tuning, verifier supervised fine-tuning, and agent policy refinement. All teacher trajectories, refreshed agent rollouts, verifier-training packages, and RL trajectories are generated exclusively from the training split.

\paragraph{Teacher trajectory generation.}
We use GPT-5.5~\cite{openai2026gpt55} as the teacher VLM and roll it out in the same evidence environment. For each $(x,y)$, the teacher receives $o_t$ and the ground-truth label $y$ as privileged supervision and acquires evidence supporting $y$. The label is used only during teacher rollout and excluded from the stored observation and all Evidence Agent inputs. At each step, the teacher produces $z_t$, containing an action and, when applicable, a non-verdict memory update, with $a_t=\operatorname{parse}(z_t)$. A trajectory is $\tau=(o_1,z_1,a_1,\ldots,o_{T_\tau},z_{T_\tau},a_{T_\tau})$. We sample two trajectories per video using different random seeds and decoding temperatures. Any rollout containing repeated, unavailable, or malformed actions, or explicit label leakage, is terminated and discarded; only trajectories with valid transitions and protocol-compliant actions are retained. Because the teacher is conditioned on $y$, terminal prediction correctness is not a filtering criterion.

\paragraph{Agent supervised fine-tuning.}
The Evidence Agent is supervised on the retained teacher trajectories. Because each response contains an action decision and, when applicable, a memory update, we apply token-level supervision to the complete response:
\begin{equation}
    \mathcal{L}_{\mathrm{SFT}}(\theta)
    =
    -\sum_{\tau}\sum_{t=1}^{T_\tau}
    \sum_{\ell=1}^{L_{\tau,t}}
    \log
    \pi_{\theta}
    \left(
    z^{*}_{\tau,t,\ell}
    \mid
    o_{\tau,t},z^{*}_{\tau,t,<\ell}
    \right),
\end{equation}
where $z_{\tau,t}^{*}$ is the teacher response. This stage teaches evidence selection, memory updating, protocol following, and stopping behavior.

\paragraph{Verifier supervised fine-tuning.}
We train the Verifier on terminal evidence packages from retained teacher trajectories and two refreshed SFT-agent rollouts per sample. Kimi K2.5~\cite{kimi2026k25} evaluates packages from both sources, and only those judged valid and consistent with the ground-truth label are retained. Let $\mathcal{D}_{\mathrm{ver}}$ denote the union of the retained teacher-terminal and agent-terminal packages, and let $y_{1:L_y}$ denote the target label tokens. With next-token distribution $p_{\phi}$, the Verifier is optimized with
\begin{equation}
    \mathcal{L}_{\mathrm{ver}}(\phi)
    =
    -\sum_{(c,\mathcal{P},y)\in\mathcal{D}_{\mathrm{ver}}}
    \sum_{\ell=1}^{L_y}
    \log p_{\phi}(y_{\ell}\mid c,\mathcal{P},y_{<\ell}).
\end{equation}
Using both package sources reduces the distribution shift between Verifier training and subsequent agent rollouts. The trained Verifier is then frozen for policy refinement.

\paragraph{Agent policy refinement.}
Finally, we refine the Evidence Agent with reinforcement learning using the frozen Verifier as the outcome evaluator. The reward combines verification outcome, evidence coverage, and interaction cost:
\begin{equation}
\begin{aligned}
\bar{R}(\tau)
&=
B(s)
+
0.80\,\mathbb{I}[s=\mathrm{correct}]Q(\tau)
-
P(s,\tau),\\
R(\tau)
&=
\operatorname{clip}\!\left(\bar{R}(\tau),-1.2,1.0\right),
\end{aligned}
\end{equation}
where $B(s)$ assigns a positive base reward to correct verification and a negative reward otherwise. For correct trajectories, $Q(\tau)$ rewards textual novelty, visual evidence, slot--channel diversity, and evidence volume, while $P(s,\tau)$ penalizes unnecessary reads, repeated or invalid actions, and forced budget termination. For incorrect trajectories, only repeated or invalid actions are penalized. The coverage bonus discourages premature stopping, especially for real videos without a single decisive contradiction. Correctness determines the primary ordering, while coverage and cost distinguish trajectories within each outcome branch.

\paragraph{Online rollout filtering.}
During GRPO refinement, all trajectories sampled for the same input are first scored by the frozen Verifier-based reward pipeline. We retain only groups with informative within-group variation according to correctness-contrast and reward-spread criteria, while degenerate groups are skipped. This filtering does not modify the reward of any individual trajectory.
For each $x$, we sample $G$ trajectories $\{\tau^{(j)}\}_{j=1}^{G}$ and compute
\begin{equation}
    \hat{A}^{(j)}
    =
    \frac{
    R(\tau^{(j)})-\mathrm{mean}_{k}R(\tau^{(k)})
    }{
    \mathrm{std}_{k}R(\tau^{(k)})
    }.
\end{equation}
The Evidence Agent generates a response
\begin{equation}
z_t^{(j)}
=
\left(
z_{t,1}^{(j)},\ldots,z_{t,L_t^{(j)}}^{(j)}
\right),
\end{equation}
at each interaction step, containing an action command and, when applicable, a memory update. We therefore define the policy ratio at the token level:
\begin{equation}
\begin{aligned}
\rho_{t,\ell}^{(j)}
=
\frac{
\pi_{\theta}\!\left(
z_{t,\ell}^{(j)}
\mid
o_t^{(j)},z_{t,<\ell}^{(j)}
\right)
}{
\pi_{\mathrm{old}}\!\left(
z_{t,\ell}^{(j)}
\mid
o_t^{(j)},z_{t,<\ell}^{(j)}
\right)
}.
\end{aligned}
\end{equation}
The asymmetric dual-clipped surrogate is
\begin{equation}
\begin{aligned}
\bar\rho_{t,\ell}^{(j)}
&=\operatorname{clip}\!\left(\rho_{t,\ell}^{(j)},1-\epsilon_{\mathrm{clip}}^{\mathrm{low}},1+\epsilon_{\mathrm{clip}}^{\mathrm{high}}\right),\\
\tilde\ell_{t,\ell}^{(j)}
&=\min\!\left(\rho_{t,\ell}^{(j)}\hat A^{(j)},\bar\rho_{t,\ell}^{(j)}\hat A^{(j)}\right),\\
\ell_{t,\ell}^{(j)}
&=\begin{cases}
\tilde\ell_{t,\ell}^{(j)}, & \hat A^{(j)}\geq0,\\
\max\!\left(\tilde\ell_{t,\ell}^{(j)},C_{\mathrm{dual}}\hat A^{(j)}\right), & \hat A^{(j)}<0.
\end{cases}
\end{aligned}
\end{equation}
The reported runs disable KL regularization; the clipping values are given in the supplementary material. The final objective is
\begin{equation}
    \mathcal{J}_{\mathrm{GRPO}}(\theta)
    =
    \mathbb{E}
    \left[
    \frac{1}{G}
    \sum_{j=1}^{G}
    \frac{1}{Z_j}
    \sum_{t=1}^{T_j}
    \sum_{\ell=1}^{L_t^{(j)}}
    \ell_{t,\ell}^{(j)}
    \right],
\end{equation}
where $Z_j=\sum_{t=1}^{T_j}L_t^{(j)}$ is the generated-token count in $\tau^{(j)}$. The same trajectory-level advantage is assigned to all tokens in a rollout, while token-count normalization prevents longer trajectories from dominating the update. We maximize $\mathcal{J}_{\mathrm{GRPO}}$ to refine the policy.

\subsection{Inference}

At inference, SIEVE alternates between the Evidence Agent and environment. The environment initializes the evidence space, coverage, history, memory, and budget, and provides a one-time slot overview. The agent then issues iterative $\mathrm{READ}$ actions or $\mathrm{STOP}$. Upon $\mathrm{STOP}$, budget exhaustion, or an invalid action, the frozen Verifier predicts
\begin{equation}
\hat{y}
=
\operatorname{parse}\!\left(g_{\phi}(c,\mathcal{P}_T)\right).
\end{equation}
This yields a binary Fake/Real label and an interpretable trail of the acquired slots and channels.

\definecolor{TableHeaderTop}{HTML}{EEF3F8}
\definecolor{TableHeaderBottom}{HTML}{FFF9F2}
\definecolor{TableOurs}{HTML}{EDF7F0}

\begin{table*}[t]
\centering
{
\small
\setlength{\tabcolsep}{3.0pt}
\renewcommand{\arraystretch}{1.08}

\begin{tabularx}{\textwidth}{@{}cl*{8}{Y}@{}}
\toprule

\multirow{2}{*}{\textbf{Paradigm}}
& \multirow{2}{*}{\textbf{Method}}
& \multicolumn{4}{c}{\cellcolor{TableHeaderTop}\textbf{FakeSV}}
& \multicolumn{4}{c}{\cellcolor{TableHeaderTop}\textbf{FakeTT}} \\

\cmidrule(lr){3-6}
\cmidrule(lr){7-10}

\rowcolor{TableHeaderBottom}
\cellcolor{white}
& \cellcolor{white}
& \textbf{ACC}
& \textbf{M-F1}
& \textbf{M-P}
& \textbf{M-R}
& \textbf{ACC}
& \textbf{M-F1}
& \textbf{M-P}
& \textbf{M-R} \\

\midrule

\multirow{5}{*}{MLLM-based}
& GPT-4o-mini\textsuperscript{*}~\cite{openai2024gpt4omini}
& 68.08 & 68.05 & 69.88 & 69.49
& 61.54 & 61.20 & 64.41 & 65.89 \\

& GPT-4.1-mini\textsuperscript{*}~\cite{openai2025gpt41}
& 70.30 & 70.25 & 70.61 & 70.87
& 49.16 & 48.54 & 62.50 & 59.70 \\

& Qwen2.5-VL\textsuperscript{*}~\cite{bai2025qwen25vl}
& 64.21 & 60.79 & 64.55 & 61.52
& 45.82 & 45.31 & 56.69 & 55.42 \\

& InternVL2.5\textsuperscript{*}~\cite{chen2024internvl25}
& 64.39 & 57.89 & 68.52 & 60.50
& 46.82 & 45.29 & 64.92 & 59.23 \\

& InternVL2.5-MPO\textsuperscript{*}~\cite{wang2024mpo}
& 65.13 & 61.07 & 66.46 & 62.12
& 43.14 & 40.84 & 61.90 & 56.23 \\

\midrule

\multirow{2}{*}{Unimodal}
& ViT\textsuperscript{*}~\cite{dosovitskiy2021vit}
& 70.85 & 70.66 & 70.64 & 70.91
& 64.88 & 62.59 & 62.54 & 63.80 \\

& BERT\textsuperscript{*}~\cite{devlin2019bert}
& 78.41 & 78.25 & 78.17 & 78.52
& 70.90 & 69.00 & 68.71 & 70.60 \\
\midrule

\multirow{4}{*}{Multimodal}
& TikTec\textsuperscript{*}~\cite{shang2021multimodal}
& 73.06 & 72.79 & 72.73 & 72.93
& 66.56 & 65.55 & 66.50 & 68.62 \\

& FANVM\textsuperscript{*}~\cite{choi2021topic}
& 79.88 & 78.91 & 80.98 & 78.42
& 71.91 & 70.85 & 71.21 & 73.90 \\

& SV-FEND\textsuperscript{*}~\cite{qi2023fakesv}
& 80.81 & 80.19 & 81.08 & 79.84
& 77.26 & 75.55 & 74.94 & 77.13 \\

& FakingRecipe\textsuperscript{*}~\cite{bu2024fakingrecipe}
& 84.69 & 84.39 & 84.57 & 84.25
& 79.26 & 77.53 & 76.86 & 78.89 \\
\midrule

\multirow{3}{*}{\shortstack{Reasoning/\\VLM-enhanced}}
& CA-FVD\textsuperscript{*}~\cite{wang2025cafvd}
& 85.79 & 85.28 & 86.57 & 84.78
& 81.61 & 80.26 & 79.50 & 82.17 \\

& ExMRD\textsuperscript{*}~\cite{hong2025following}
& 86.90 & 86.52 & 87.31 & 86.13
& 84.28 & 83.13 & 82.27 & 85.19 \\

& FakeSV-VLM\textsuperscript{*}~\cite{wang2025fakesvvlm}
& \underline{90.22}
& \underline{89.97}
& \underline{90.55}
& \underline{89.64}
& \underline{89.30}
& \underline{87.98}
& \underline{87.80}
& \underline{88.17} \\
\midrule

\rowcolor{TableOurs}
\textbf{Ours}
& \textbf{SIEVE}
& \textbf{92.62}
& \textbf{92.41}
& \textbf{93.25}
& \textbf{91.96}
& \textbf{91.64}
& \textbf{90.72}
& \textbf{90.12}
& \textbf{91.45} \\

\bottomrule
\end{tabularx}
}

\caption{Performance comparison on FakeSV and FakeTT. Best results are shown in \textbf{bold}, and second-best results are underlined. SIEVE results are averaged over three independent runs. Results marked with \textsuperscript{*} are taken from FakeSV-VLM~\cite{wang2025fakesvvlm}.}
\label{tab:main_results}
\vspace{-8pt}
\end{table*}

\section{Experiments}

\noindent\textbf{Datasets.}
We evaluate SIEVE on FakeSV and FakeTT, with out-of-domain testing on FakeVV. FakeSV is a Chinese benchmark from Douyin and Kuaishou~\cite{qi2023fakesv}, while FakeTT is an English TikTok benchmark introduced by FakingRecipe~\cite{bu2024fakingrecipe}. Both provide Event and Description fields, and FakeSV additionally includes comments. SIEVE uses only Description as the initial claim context and excludes all other fields. Following prior work, we adopt a temporal 70\%/15\%/15\% train/validation/test split. For out-of-domain evaluation, the FakeTT-trained checkpoint is directly applied to FakeVV~\cite{zhang2025factr1}, where each query pairs a source video with an authentic or entity-manipulated title. SIEVE uses the title as the claim context and retrieves evidence only from the paired video, without using any FakeVV data for training, validation, model selection, or prompt adaptation.

\noindent\textbf{Baselines.}
For FakeSV and FakeTT, we compare SIEVE with four baseline groups: MLLMs, including GPT-4o-mini, GPT-4.1-mini, Qwen2.5-VL, InternVL2.5, and InternVL2.5-MPO~\cite{openai2024gpt4omini,openai2025gpt41,bai2025qwen25vl,chen2024internvl25,wang2024mpo}; unimodal models, ViT and BERT~\cite{dosovitskiy2021vit,devlin2019bert}; task-specific multimodal methods, TikTec, FANVM, SV-FEND, and FakingRecipe~\cite{shang2021multimodal,choi2021topic,qi2023fakesv,bu2024fakingrecipe}; and reasoning- or VLM-enhanced methods, CA-FVD, ExMRD, and FakeSV-VLM~\cite{wang2025cafvd,hong2025following,wang2025fakesvvlm}. All in-domain baseline results are taken from FakeSV-VLM~\cite{wang2025fakesvvlm}. For FakeVV, we compare against GPT-4o, DeepSeek-R1, Qwen2.5-VL-7B, QVQ-72B-Preview, InternVL2.5-8B, Qwen3-VL-8B-Instruct, and GPT-5.5~\cite{openai2024gpt4o,guo2025deepseekr1,bai2025qwen25vl,qwen2024qvq72bpreview,chen2024internvl25,bai2025qwen3vl,openai2026gpt55}. Results for the first five are reported by Fact-R1~\cite{zhang2025factr1}, whereas the last two are evaluated by us under the same protocol. Fact-R1 is excluded because its reported model is trained on FakeVV, violating our no-FakeVV-training setting.

\noindent\textbf{Training Details.}
The agent and verifier use Qwen3-VL-8B-Instruct and Qwen3-VL-2B-Instruct~\cite{bai2025qwen3vl}, respectively, with LoRA adapters. The agent undergoes one epoch of SFT on teacher trajectories, followed by 100 GRPO steps with group size $n=8$, a read budget of 24, and a 32K-token context window. The verifier is trained for two epochs with up to 24 images per example. Learning rates are $1\times10^{-4}$ for agent SFT and verifier training and $5\times10^{-7}$ for RL, with per-device batch size 1 and gradient accumulation 4. We report the mean over three independent runs with seeds 42, 2027, and 3407, using a fixed test-time decoding temperature of 0.7. All experiments use 8 NVIDIA A800 80GB GPUs with bfloat16 precision.

\subsection{Main Results}

Table~\ref{tab:main_results} reports the main results on FakeSV and FakeTT. SIEVE achieves the best performance across all metrics on both benchmarks. On FakeSV, it obtains 92.62\% ACC and 92.41\% Macro-F1, outperforming the strongest prior method, FakeSV-VLM, by 2.40 and 2.44 points, respectively. On FakeTT, SIEVE improves ACC from 89.30\% to 91.64\% and Macro-F1 from 87.98\% to 90.72\%, yielding gains of 2.34 and 2.74 points. Several trends can be observed. Under the evaluated settings, general-purpose MLLMs perform worse than task-specific methods, indicating that direct prompting alone remains challenging for fine-grained video misinformation detection. Stronger multimodal detectors generally outperform unimodal baselines, although early multimodal fusion methods do not always surpass a strong text-only model, suggesting that adding modalities does not by itself guarantee better performance. Reasoning- and VLM-enhanced methods further improve performance, yet they still mainly operate on preconstructed global inputs. In contrast, SIEVE explicitly separates evidence acquisition from veracity judgment and exposes the verifier only to selected OCR, ASR, and visual evidence. The consistent gains on both Chinese and English benchmarks demonstrate the effectiveness of the proposed acquisition-and-verification framework under the evaluated settings. Together with the evidence-budget analysis, these results indicate that compact, claim-relevant evidence can support strong verification performance without exposing all constructed evidence to the verifier.


\definecolor{TableHeaderTop}{HTML}{EEF3F8}
\definecolor{TableHeaderBottom}{HTML}{FFF9F2}
\definecolor{TableOurs}{HTML}{EDF7F0}

\begin{table}[t]
\centering

{
\small
\setlength{\tabcolsep}{2.2pt}
\renewcommand{\arraystretch}{1.08}

\begin{tabularx}{\columnwidth}{
    @{}
    >{\raggedright\arraybackslash}p{0.19\columnwidth}
    X
    cccc
    @{}
}
\toprule

\multirow[c]{2}{*}{
    \shortstack[c]{\textbf{Ablation}\\\textbf{Type}}
}
&
\multirow[c]{2}{*}{\textbf{ \ \ \ \ \ \ \ Setting}}
&
\multicolumn{2}{
    >{\columncolor{TableHeaderTop}}c
}{
    \rule{0pt}{3.0ex}\textbf{FakeSV}
}
&
\multicolumn{2}{
    >{\columncolor{TableHeaderTop}}c
}{
    \rule{0pt}{3.0ex}\textbf{FakeTT}
}
\\

\cmidrule(lr){3-4}
\cmidrule(lr){5-6}

&
&
\cellcolor{TableHeaderBottom}
\rule{0pt}{3.0ex}\textbf{ACC}
&
\cellcolor{TableHeaderBottom}
\textbf{M-F1}
&
\cellcolor{TableHeaderBottom}
\textbf{ACC}
&
\cellcolor{TableHeaderBottom}
\textbf{M-F1}
\\

\midrule

\multirow[c]{3}{*}{
    \shortstack[c]{Evidence\\Input}
}
&
Description-only
&
77.12
&
73.92
&
66.22
&
65.72
\\

&
Same-video random
&
61.99
&
49.20
&
55.85
&
55.43
\\

&
Exhaustive evidence
&
92.25
&
91.97
&
89.63
&
87.80
\\

\midrule

\multirow[c]{4}{*}{
    \shortstack[c]{Agent\\Training}
}
&
Base agent
&
88.56
&
87.91
&
84.95
&
84.23
\\

&
Agent SFT only
&
91.51
&
91.13
&
89.97
&
89.13
\\

&
w/o rollout filtering
&
91.70
&
91.33
&
90.64 
& 
89.58
\\

&
w/o evidence memory
&
91.88
&
91.63
&
90.97
&
89.78
\\

\midrule

\rowcolor{TableOurs}
Full Model
&
\textbf{Full SIEVE}
&
\textbf{92.62}
&
\textbf{92.41}
&
\textbf{91.64}
&
\textbf{90.72}
\\

\bottomrule
\end{tabularx}
}

\caption{Ablation study on FakeSV and FakeTT. Exhaustive evidence trains a separate verifier on all OCR/ASR evidence and up to 24 uniformly sampled images; all other variants share a fixed verifier.}
\label{tab:ablation_both}
\vspace{-8pt}
\end{table}

\subsection{Ablation Study}

To assess each component, we evaluate seven variants: (1) \textbf{Description-only}, using only the Description field; (2) \textbf{Same-video random evidence}, replacing the selected package with an equally sized random package from the same video; (3) \textbf{Exhaustive evidence}, training and evaluating a separate Verifier with all constructed OCR and ASR evidence and up to 24 images uniformly sampled from the full visual evidence space; (4) \textbf{Base agent}, removing agent-side SFT and RL; (5) \textbf{Agent SFT only}, removing RL; (6) \textbf{w/o rollout filtering}, disabling group-level filtering during RL; and (7) \textbf{w/o evidence memory}, removing cross-step evidence memory. Except for Exhaustive evidence, the Verifier is fixed across all variants. As shown in Table~\ref{tab:ablation_both}, Description-only confirms the complementary value of acquired evidence, while same-video random evidence shows that indiscriminate selection is insufficient. Despite being separately trained and evaluated with comprehensive textual evidence and uniformly sampled visual evidence, Exhaustive evidence still underperforms Full SIEVE on both datasets, particularly by 2.92 M-F1 points on FakeTT. This indicates that broad evidence coverage alone cannot replace targeted acquisition: redundant or weakly informative content may dilute decisive clues, whereas SIEVE constructs a compact and verification-effective evidence package. Agent SFT substantially improves over the base agent, validating trajectory-level supervision. Removing rollout filtering or evidence memory further degrades performance, supporting their roles in policy refinement and coherent acquisition. Overall, Full SIEVE performs best on both datasets, demonstrating the joint contribution of targeted acquisition and agent training.

\subsection{Can Sparse Evidence Suffice?}
\label{sec:evidence_budget}

\begin{figure}[t]
    \centering
    \includegraphics[width=\columnwidth]{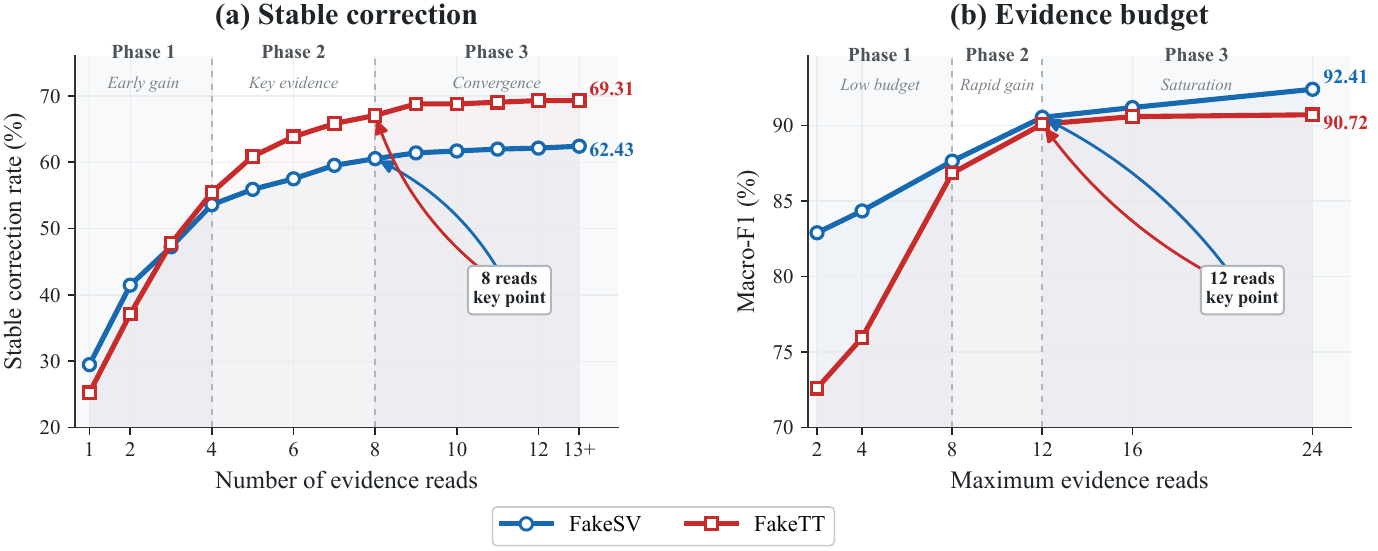}
    \caption{Evidence sufficiency on FakeSV and FakeTT.
    (a) Stable correction of Description-only errors.
    (b) Macro-F1 under different inference-time evidence budgets.}
    \label{fig:evidence_budget}
    \vspace{-8pt}
\end{figure}

Figure~\ref{fig:evidence_budget} investigates whether reliable
verification can be achieved from only a small subset of the
available evidence. In Figure~\ref{fig:evidence_budget}(a), we focus
on samples misclassified under the Description-only setting and
evaluate the frozen Verifier after each legal read. A sample is
counted at step $k$ only if its prediction remains correct for all
subsequent evidence prefixes. Within four reads, 53.61\% and
55.45\% of the initial errors are stably corrected on FakeSV and
FakeTT, respectively. The rates reach 60.55\% and 67.08\% within
eight reads, accounting for approximately 97\% of all eventually
corrected cases on both datasets. This shows that, for a substantial
fraction of samples requiring video evidence, only a few acquired
clues are sufficient to establish a stable correct prediction. Figure~\ref{fig:evidence_budget}(b) examines whether such compact
evidence also preserves overall performance. Increasing the
inference-time cap from 2 to 12 reads improves Macro-F1 from
82.90\% to 90.55\% on FakeSV and from 72.60\% to 90.10\% on
FakeTT, whereas extending the cap from 12 to 24 yields only 1.86
and 0.62 additional points. Together, the rapid stable correction
and early performance saturation provide direct empirical support
for our central hypothesis: sparse, claim-relevant evidence can
suffice for reliable video misinformation verification under the
considered evidence space and Verifier interface.



\begin{figure}[t]
    \centering
    \includegraphics[width=\linewidth]{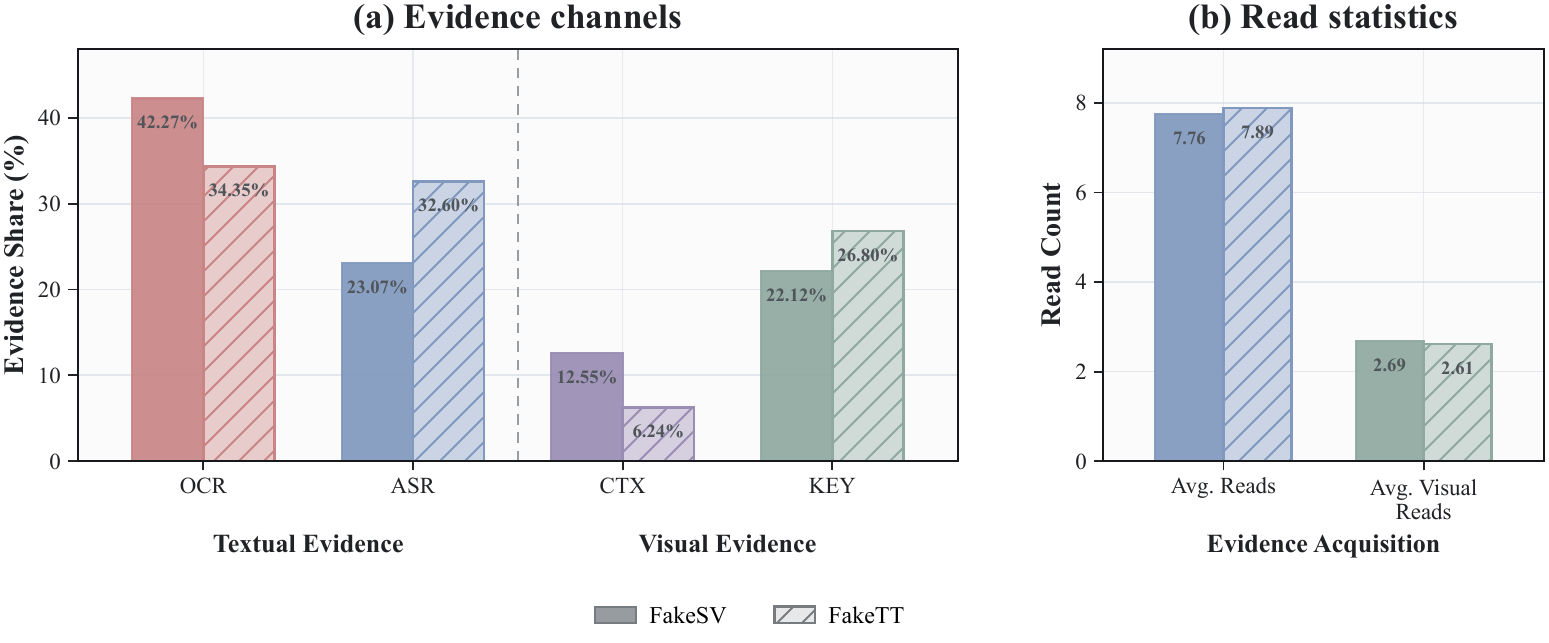}
    \caption{Evidence acquisition statistics on the test sets. The left panel shows the distribution of legal reads across evidence channels, and the right reports average total and visual reads.}
    \label{fig:evidence_acquisition}.
    \vspace{-8pt}
\end{figure}

\subsection{Evidence Acquisition Analysis}
\label{sec:evidence_acquisition}

Figure~\ref{fig:evidence_acquisition} summarizes the evidence acquisition behavior of SIEVE under the 24-read inference cap. The agent uses fewer than eight legal read actions per instance on average: 7.76 on FakeSV and 7.89 on FakeTT. Under the current action and Verifier interface, it therefore constructs a compact evidence package rather than reading every available slot--channel pair. Textual evidence is selected most frequently. On FakeSV, OCR accounts for 42.27\% of legal reads and ASR for 23.07\%. On FakeTT, OCR and ASR are more balanced at 34.35\% and 32.60\%, respectively. This pattern suggests that the learned policy adapts its channel preference to dataset-specific evidence distributions. Visual evidence is used more selectively. The agent averages 2.69 and 2.61 visual reads on FakeSV and FakeTT, respectively, and selects KEY frames more often than CTX frames. This selection pattern is consistent with the learned policy favoring visually distinctive KEY frames over generic contextual CTX frames. Overall, the results show that strong verification performance is obtained from compact, complementary textual and visual evidence packages within the considered evidence budget.

\subsection{Out-of-Domain Generalization}
Table~\ref{tab:ood_fakevv} evaluates cross-dataset generalization by directly applying the FakeTT-trained SIEVE checkpoint to FakeVV, without using FakeVV for training, validation, model selection, or prompt adaptation. SIEVE achieves 73.20\% ACC and 74.91\% F1, outperforming all evaluated baselines under this no-FakeVV-training protocol. Its recall of 80.00\% and precision of 70.42\% indicate that SIEVE identifies most fake cases across domains, although this relatively high sensitivity is accompanied by some false-positive predictions. We exclude Fact-R1 itself because its reported model is trained on FakeVV and is therefore not directly comparable in this out-of-domain setting.

\definecolor{TableHeaderTop}{HTML}{EEF1F6}  
\definecolor{TableOurs}{HTML}{EDF4F2}       

\begin{table}[t]
\centering

{
\small
\setlength{\tabcolsep}{2.6pt}
\renewcommand{\arraystretch}{1.08}

\begin{tabularx}{\columnwidth}{
@{}
>{\raggedright\arraybackslash}X
cccc
@{}
}
\toprule

\rowcolor{TableHeaderTop}
\textbf{Method}
& \textbf{ACC}
& \textbf{Prec.}
& \textbf{Rec.}
& \textbf{F1} \\

\midrule

GPT-4o\textsuperscript{\(\dagger\)}
~\cite{openai2024gpt4o}
& 56.0 & 60.4 & 35.0 & 44.3 \\

DeepSeek-R1\textsuperscript{\(\dagger\)}
~\cite{guo2025deepseekr1}
& 53.5 & 58.1 & 25.2 & 35.1 \\

Qwen2.5-VL-7B\textsuperscript{\(\dagger\)}
~\cite{bai2025qwen25vl}
& 52.9 & 51.1 & 51.1 & 51.1 \\

QVQ-72B-Preview\textsuperscript{\(\dagger\)}
~\cite{qwen2024qvq72bpreview}
& 53.5 & 52.6 & 52.6 & 52.6 \\

InternVL2.5-8B\textsuperscript{\(\dagger\)}
~\cite{chen2024internvl25}
& 53.5 & 58.5 & 24.0 & 34.0 \\

Qwen3-VL-8B~\cite{bai2025qwen3vl}
& 67.3 & 66.3 & 72.5 & 69.2 \\

GPT-5.5~\cite{openai2026gpt55}
& 67.8 & 66.6 & 73.3 & 69.8 \\

\midrule

\rowcolor{TableOurs}
\textbf{SIEVE (FakeTT ckpt.)}
& \textbf{73.2}
& \textbf{70.4}
& \textbf{80.0}
& \textbf{74.9} \\

\bottomrule
\end{tabularx}
}

\caption{Out-of-domain performance on FakeVV under the no-FakeVV-training setting. \emph{Fake} is the positive class. Best results are in \textbf{bold}; \textsuperscript{\(\dagger\)} marks results reported by Fact-R1~\cite{zhang2025factr1}.}
\label{tab:ood_fakevv}
\end{table}



\bibliography{aaai2027}

\clearpage
\appendix

\section{Dataset Splits and Statistics}
Table~\ref{tab:dataset_statistics} summarizes the datasets used in our experiments. FakeSV and FakeTT are chronologically partitioned into training, validation, and test sets, while FakeVV is used only for out-of-domain evaluation with its balanced official test set. The class distributions of fake and real samples are also reported for completeness.

\begin{table}[t]
\centering

{
\small
\setlength{\tabcolsep}{3.2pt}
\renewcommand{\arraystretch}{1.10}

\begin{tabular}{@{}lccccc@{}}
\toprule
\textbf{Dataset}
& \textbf{Total}
& \textbf{Overall F/R}
& \textbf{Train}
& \textbf{Val.}
& \textbf{Test} \\
\midrule

FakeSV
& 3,624
& 1,810/1,814
& 2,536
& 546
& 542 \\

FakeTT
& 1,991
& 1,172/819
& 1,393
& 299
& 299 \\

FakeVV
& 102,000
& 51,000/51,000
& 100,000
& --
& 2,000 \\

\bottomrule
\end{tabular}
}

\caption{Dataset statistics. FakeSV and FakeTT use chronological train/validation/test splits of approximately 70\%/15\%/15\%. FakeVV contains 102,000 video--text pairs; only its balanced test set of 2,000 queries from 997 unique source videos is used for out-of-domain evaluation. F/R denotes fake/real sample counts.}
\label{tab:dataset_statistics}
\end{table}

\section{Evidence Space Construction: Implementation Details}
\label{app:evidence-space}

\paragraph{Shot detection and slot refinement.}
Shot boundaries are detected with TransNetV2~\cite{soucek2020transnetv2}. Duration-based refinement then proceeds in two passes. First, shots shorter than $0.8$~s are merged into the temporally shorter of their two neighbors. Second, shots longer than $6$~s are subdivided using motion-equalized splitting based on frame differences, with uniform-duration splitting used when motion estimates are unavailable. Depending on duration, a long shot is divided into two segments up to $12$~s, three segments up to $18$~s, and four segments beyond $18$~s.

If the number of refined slots still exceeds $N_{\max}=10$, we iteratively merge the adjacent pair with the lowest \emph{motion density}, defined as the combined frame-difference motion divided by the combined duration. This preferentially merges low-information static segments while preserving short but dynamic slots. When per-frame motion cannot be computed, the shortest adjacent pair is merged instead.

\paragraph{OCR channel.}
On-screen text is extracted with PaddleOCR~\cite{du2020ppocr} at 2 FPS, using a minimum box duration of $0.5$~s and a minimum area ratio of $1.5\%$. Deduplicated OCR items within each slot are ranked by
\[
\operatorname{imp}
=
0.25\hat d+0.30\hat a+0.20c+0.15\hat s+0.10\hat\ell,
\]
where $\hat d=\min(d/5\text{ s},1)$, $\hat a=\min(a/0.1,1)$, $c$ is the mean OCR confidence, $\hat s$ is the temporal-stability score of the text box across consecutive detections, and $\hat\ell=\min(\ell/20,1)$ for character length $\ell$. The OCR channel exposes the deduplicated text items in descending importance order.

\paragraph{ASR channel.}
Full-video transcripts are produced by Whisper large-v3~\cite{radford2023whisper} with word-level timestamps. Each transcript segment is matched to a slot using a 1-s context-extended window $[t_{\mathrm{start}}-1,t_{\mathrm{end}}+1]$. A segment is designated \emph{primary} when its midpoint lies within $[t_{\mathrm{start}},t_{\mathrm{end}}]$ and \emph{boundary} when it overlaps only the extended window. The ASR channel exposes only the concatenated primary text, assigning each segment to exactly one slot; boundary context is retained as metadata but is not shown to the agent.

\paragraph{Visual channels: CTX and KEY.}
Candidate frames are encoded with DINOv2 ViT-S/14~\cite{oquab2024dinov2} using the 384-dimensional, $\ell_2$-normalized CLS token, such that cosine similarity reduces to a dot product. CTX contains three representative frames: the slot is divided into three equal-duration bins, and one frame per bin is selected by jointly considering sharpness, measured by Laplacian variance, and dissimilarity to previously selected CTX frames.

KEY contains up to two complementary frames:
\begin{itemize}
    \item \texttt{key\_diff} is the candidate with the lowest maximum cosine similarity to the three CTX frames, with sharpness used as a tie-breaker. If all candidates have similarity above $0.97$, the least-similar candidate is retained as a fallback.
    \item \texttt{key\_ocr} is the sharpest frame within the temporal window of the OCR item with the highest $\operatorname{imp}$ score.
\end{itemize}
If \texttt{key\_diff} and \texttt{key\_ocr} coincide, or if one is unavailable, KEY contains a single frame.

\paragraph{Coverage matrix and legal actions.}
The coverage matrix $\mathbf{C}_t\in\{\mathtt{-},\mathtt{x},\mathtt{v}\}^{|\mathcal{S}|\times|\mathcal{M}|}$ is initialized to $\mathtt{x}$ for non-empty channels and $\mathtt{-}$ otherwise; $\mathtt{v}$ denotes a read channel. Let $n_t^{\mathrm{img}}$ be the number of accumulated CTX/KEY images and $\kappa_{i,m}$ the images returned by a READ, with $\kappa_{i,m}=0$ for OCR and ASR. With $B_{\mathrm{img}}=24$, the legal read set is
\[
\begin{aligned}
\mathcal{A}^{\mathrm{read}}_t
=
\bigl\{\,
&\mathrm{READ}(\mathrm{slot}_{\mathrm{sid}},
                \{\mathrm{channel}\})
\\[-1pt]
&\;\bigm|\;
\mathbf{C}_t(\mathrm{sid},\mathrm{channel})=\mathtt{x},
\quad b^{\mathrm{read}}_t>0,
\\[-1pt]
&\phantom{\;\bigm|\;}
n_t^{\mathrm{img}}
+\kappa_{\mathrm{sid},\mathrm{channel}}
\le B_{\mathrm{img}}
\,\bigr\}.
\end{aligned}
\]
The visual READs excluded by this constraint are explicitly listed in \texttt{<budget\_state>} for the agent.
Each legal READ consumes one unit of the overall read budget and changes its coverage status from $\mathtt{x}$ to $\mathtt{v}$. Repeated, unavailable, or malformed actions immediately terminate the current acquisition rollout, are excluded from the evidence package, and receive the behavioral penalty used during reinforcement learning.

\section{Training and Optimization Details}
\label{app:training-details}

\subsection{Reward Design Details}
\label{app:reward}

The trajectory reward combines verification outcome, evidence coverage, and interaction cost. Let $s(\tau)$ denote the terminal status of trajectory $\tau$. As introduced in the main text, the reward is
\begin{equation}
\begin{aligned}
\bar{R}(\tau)
&=
B(s)
+
0.80\,\mathbb{I}[s=\mathrm{correct}]Q(\tau)
-
P(s,\tau),\\
R(\tau)
&=
\operatorname{clip}\!\left(\bar{R}(\tau),-1.2,1.0\right).
\end{aligned}
\end{equation}

\paragraph{Verification outcome.}
The base reward depends on the terminal verification outcome:
\begin{equation}
B(s)=
\begin{cases}
0.20, & s=\mathrm{correct},\\
-1.00, & s=\mathrm{incorrect},\\
-1.00, & s=\mathrm{no\ valid\ evidence},\\
-1.00, & s=\mathrm{malformed}.
\end{cases}
\end{equation}
Here, \textit{no valid evidence} denotes termination without any legally acquired evidence, while \textit{malformed} denotes a terminal Verifier output that cannot be parsed into a valid binary prediction.

\paragraph{Evidence coverage.}
For a correct trajectory, the evidence-coverage score is
\begin{equation}
\label{eq15}
\begin{aligned}
Q(\tau)
={}&
0.45q_{\mathrm{text}}
+0.25q_{\mathrm{vis}}
+0.20q_{\mathrm{div}}
+0.10q_{\mathrm{vol}},\\
q_{\mathrm{text}}
={}&
\operatorname{clip}
\left(
\frac{n_{\mathrm{novel}}}{180},0,1
\right),\\
q_{\mathrm{vis}}
={}&
\operatorname{clip}
\left(
\max\left(\frac{n_v}{2},
          \frac{n_{\mathrm{img}}}{4}\right),
0,1
\right),\\
q_{\mathrm{div}}
={}&
\operatorname{clip}
\left(
0.5\frac{n_{\mathrm{slot}}}{2}
+
0.5\frac{n_{\mathrm{ch}}}{2},
0,1
\right),\\
q_{\mathrm{vol}}
={}&
\operatorname{clip}
\left(
\frac{n_r}{3},0,1
\right).
\end{aligned}
\end{equation}
Here, $n_{\mathrm{novel}}$ is the number of OCR or ASR characters not already present in the initial claim context; $n_v$ and $n_{\mathrm{img}}$ denote the numbers of visual reads and acquired images; $n_{\mathrm{slot}}$ and $n_{\mathrm{ch}}$ denote the numbers of distinct slots and evidence channels covered; and $n_r$ is the number of valid reads. The coverage bonus is applied only to correct trajectories and discourages premature stopping.

\paragraph{Interaction penalties.}
We define the efficiency, behavioral, and forced-termination penalties as
\begin{equation}
\begin{aligned}
P_{\mathrm{eff}}
&=
\min\left(
0.15,\,
0.01n_r+0.005n_v
\right),\\
P_{\mathrm{beh}}
&=
\min\left(
0.30,\,
0.15n_{\mathrm{inv}}
\right)
+
\min\left(
0.20,\,
0.08n_{\mathrm{rep}}
\right),\\
P_{\mathrm{stop}}
&=
0.03\,
\mathbb{I}[
\text{forced termination at budget exhaustion}
],
\end{aligned}
\end{equation}
where $n_{\mathrm{inv}}$ counts unavailable or malformed actions,
excluding repeated reads, while $n_{\mathrm{rep}}$ counts repeated
read attempts. The two action categories are mutually exclusive.

The branch-dependent penalty is
\begin{equation}
P(s,\tau)=
\begin{cases}
P_{\mathrm{eff}}+P_{\mathrm{beh}}+P_{\mathrm{stop}},
& s=\mathrm{correct},\\
P_{\mathrm{beh}},
& s=\mathrm{incorrect},\\
\min(0.30,0.15n_{\mathrm{inv}}),
& s=\mathrm{no\ valid\ evidence},\\
0,
& s=\mathrm{malformed}.
\end{cases}
\end{equation}

This design makes terminal correctness the primary reward signal. Correct trajectories receive additional credit for acquiring complementary textual and visual evidence, while excessive reads, repeated actions, invalid actions, and budget-exhausted termination are discouraged. All final rewards are clipped to $[-1.2,1.0]$ for stable policy optimization.

\subsection{GRPO Optimization Details}
\label{app:grpo}

For completeness, we provide the sampled-token KL estimator supported by the GRPO implementation. Let
$u_{t,\ell}^{(j)}=(o_t^{(j)},z_{t,<\ell}^{(j)})$. We define
\begin{equation}
\begin{aligned}
    \Delta_{t,\ell}^{(j)}
    &=
    \log \pi_{\theta}\!\left(
    z_{t,\ell}^{(j)}\mid u_{t,\ell}^{(j)}
    \right)
    -
    \log \pi_{\mathrm{ref}}\!\left(
    z_{t,\ell}^{(j)}\mid u_{t,\ell}^{(j)}
    \right),\\
    \widehat{D}_{\mathrm{KL}}^{(j,t,\ell)}
    &=
    \exp\!\left(-\Delta_{t,\ell}^{(j)}\right)
    +\Delta_{t,\ell}^{(j)}-1.
\end{aligned}
\end{equation}
Here, $\pi_{\mathrm{ref}}$ is the fixed reference policy. In our formal runs the KL penalty is disabled ($\beta=0$, i.e.\ \texttt{disable\_kl=true}, \texttt{kl\_coef=0.0}); we retain the estimator above for completeness since it is the framework's default mechanism and may be enabled in future ablations.

The policy-ratio clipping range follows the asymmetric PPO/DAPO convention:
\begin{equation}
\epsilon_{\mathrm{clip}}^{\mathrm{low}} = 0.20, \qquad
\epsilon_{\mathrm{clip}}^{\mathrm{high}} = 0.30, \qquad
C_{\mathrm{dual}} = 3.0,
\end{equation}
where $C_{\mathrm{dual}}$ is the dual-clip constant applied when the advantage is strongly
negative. The advantage estimator is GRPO (group-relative, no value network): for a group of
$n=8$ rollouts sharing a prompt, $\hat{A}^{(j)} = (R^{(j)} - \mathrm{mean}(R))/\mathrm{std}(R)$.

\paragraph{Online rollout filtering.}
Before computing advantages, we discard low-information groups using a hybrid criterion
operating on three per-rollout reward-manager outputs: the overall reward $R^{(j)}$, the binary
correctness $a^{(j)}\in\{0,1\}$, and the evidence-quality score $q^{(j)}=Q(\tau^{(j)})$ (as
defined in Eq.~\ref{eq15}). For a group $g$ of rollouts sharing a prompt, define the
\emph{correctness-contrast} indicator
\begin{equation}
\begin{aligned}
\mathrm{mixed}(g) ={}&
\mathbb{I}\!\left[\max_j a^{(j)} > 0\right]
\\&
\wedge\
\mathbb{I}\!\left[\min_j a^{(j)} \le 0\right],
\end{aligned}
\end{equation}
i.e.\ whether the group contains both correct and incorrect trajectories, and the
\emph{reward-spread} indicator
\begin{equation}
\begin{aligned}
\mathrm{spread}(g) ={}&
\mathbb{I}\!\left[\max_j R^{(j)} - \min_j R^{(j)} \ge \rho_{\mathrm{range}}\right]
\\&
\vee\
\mathbb{I}\!\left[\mathrm{std}_j\!\left(R^{(j)}\right) \ge \rho_{\mathrm{std}}\right],
\end{aligned}
\end{equation}
i.e.\ whether the group's overall-reward values are sufficiently dispersed, with
$\rho_{\mathrm{range}}=0.04$ and $\rho_{\mathrm{std}}=0.015$. Writing
$j^\star=\arg\max_j R^{(j)}$ for the top-scoring rollout in the group, define the
all-correct admissibility indicator
\begin{equation}
\begin{aligned}
\mathrm{good}(g) ={}&
\mathbb{I}\!\left[\min_j a^{(j)} > 0\right]
\wedge
\mathrm{spread}(g)
\\&
\wedge\
\mathbb{I}\!\left[q^{(j^\star)} \ge \tau_{\mathrm{evi}}\right]
\\&
\wedge\
\mathbb{I}\!\left[\max_j R^{(j)} > 0.01\right]
\\&
\wedge\
\mathbb{I}\!\left[\min_j R^{(j)} < 0.99\right],
\end{aligned}
\end{equation}
with evidence floor $\tau_{\mathrm{evi}}=0.40$. A group is kept iff
\begin{equation}
\mathrm{keep}(g) = \mathrm{mixed}(g) \ \vee\ \mathrm{good}(g).
\end{equation}
That is, groups with a correctness contrast are
always kept regardless of reward dispersion; all-correct groups are kept only if their
top-scoring trajectory has adequate evidence support and the group is not reward-degenerate;
all-incorrect groups are always discarded, so that an unresolved or wrong-verifier outcome is
never learned as a relative improvement over another wrong outcome.

\paragraph{Teacher decoding.}
Teacher trajectories for SFT supervision are sampled at two fixed decoding temperatures,
$T\in\{0.2, 0.7\}$, with one trajectory generated per (video, temperature) pair; the two
temperature pools are merged and deduplicated before SFT data construction.

\paragraph{LoRA configuration.}
Both the Agent and the Verifier use LoRA with rank $r=32$,
scaling $\alpha=64$, and dropout $0.05$ during both the SFT and
RL stages. The RL stage is initialized from the SFT adapter and
retains a dropout value of $0.05$. LoRA is applied to
\texttt{q\_proj}, \texttt{k\_proj}, \texttt{v\_proj},
\texttt{o\_proj}, \texttt{gate\_proj}, \texttt{up\_proj},
and \texttt{down\_proj}. The RL actor's LoRA adapter is initialized
from the Agent SFT checkpoint specified by
\texttt{init\_lora\_path}, while the Verifier's LoRA adapter remains
frozen throughout RL.

\paragraph{Optimizer.}
The RL actor is optimized with AdamW, learning rate $5\times10^{-7}$, weight decay
$10^{-2}$, and no warmup (\texttt{lr\_warmup\_ratio=0.0}). Agent/Verifier SFT instead uses learning rate $10^{-4}$ with a cosine schedule and a
$3\%$ linear warmup. Rollout uses temperature $1.0$, top-$p$ $1.0$, group size $n=8$ during
training, and temperature $0.7$/top-$p$ $0.9$/$n=1$ for validation.

\section{Reward Ablation Study}
\label{app:reward_ablation}
Table~\ref{tab:reward_ablation} examines how different reward components affect verification performance and evidence-acquisition efficiency. Using only the terminal outcome provides limited gains over SFT, indicating that correctness supervision alone is insufficient to learn an effective evidence-seeking policy. Removing the coverage reward reduces the average number of reads to 6.94 on FakeSV and 7.03 on FakeTT, but consistently degrades ACC and Macro-F1. It also lowers Real-class recall relative to the full reward, suggesting that coverage guidance helps prevent premature stopping when no single decisive contradiction is available. In contrast, removing interaction penalties substantially increases the average number of reads to 10.84 and 11.21, while still underperforming the full model. This shows that unrestricted evidence acquisition introduces redundant or less informative observations rather than reliably improving verification. The complete reward achieves the best ACC and Macro-F1 on both datasets with fewer than eight reads on average. Although some ablations obtain higher Real-class recall on FakeTT, their lower balanced metrics indicate a class-specific trade-off rather than superior overall verification. Overall, the coverage reward and interaction penalties play complementary roles: the former encourages sufficient evidence acquisition, whereas the latter discourages unnecessary exploration, jointly producing a more accurate and efficient policy.

\definecolor{TableHeaderTop}{HTML}{EEF3F8}
\definecolor{TableOurs}{HTML}{EDF7F0}

\begin{table}[t]
\centering

{
\small
\setlength{\tabcolsep}{2.0pt}
\renewcommand{\arraystretch}{1.08}

\begin{tabularx}{\columnwidth}{@{}lXcccc@{}}
\toprule

\rowcolor{TableHeaderTop}
\textbf{Dataset}
& \textbf{Setting}
& \textbf{ACC $\uparrow$}
& \textbf{M-F1 $\uparrow$}
& \shortstack{\textbf{R-Rec.}\\$\uparrow$}
& \shortstack{\textbf{Avg. Reads}\\$\downarrow$} \\

\midrule

\multirow{5}{*}{FakeSV}
& SFT only
& 91.51 & 91.13 & 80.67 & 8.34 \\

& Outcome only
& 91.51 & 91.19 & 82.35 & 8.62 \\

& w/o coverage
& 91.88 & 91.60 & 83.61 & \textbf{6.94} \\

& w/o penalties
& 92.25 & 92.01 & 85.29 & 10.84 \\

& \cellcolor{TableOurs}\textbf{Full}
& \cellcolor{TableOurs}\textbf{92.62}
& \cellcolor{TableOurs}\textbf{92.41}
& \cellcolor{TableOurs}\textbf{86.55}
& \cellcolor{TableOurs}\underline{7.76} \\

\midrule

\multirow{5}{*}{FakeTT}
& SFT only
& 89.97 & 89.13 & \textbf{93.94} & 8.51 \\

& Outcome only
& 90.30 & 89.43 & \underline{92.93} & 8.78 \\

& w/o coverage
& 90.64 & 89.63 & 89.90 & \textbf{7.03} \\

& w/o penalties
& 91.30 & 90.37 & 90.91 & 11.21 \\

& \cellcolor{TableOurs}\textbf{Full}
& \cellcolor{TableOurs}\textbf{91.64}
& \cellcolor{TableOurs}\textbf{90.72}
& \cellcolor{TableOurs}90.91
& \cellcolor{TableOurs}\underline{7.89} \\

\bottomrule
\end{tabularx}
}

\caption{Reward-component ablation on FakeSV and FakeTT. R-Rec denotes recall for the Real class.}
\label{tab:reward_ablation}
\end{table}

\section{Prompts}

\begin{figure}[t]
\centering
\includegraphics[width=\columnwidth]{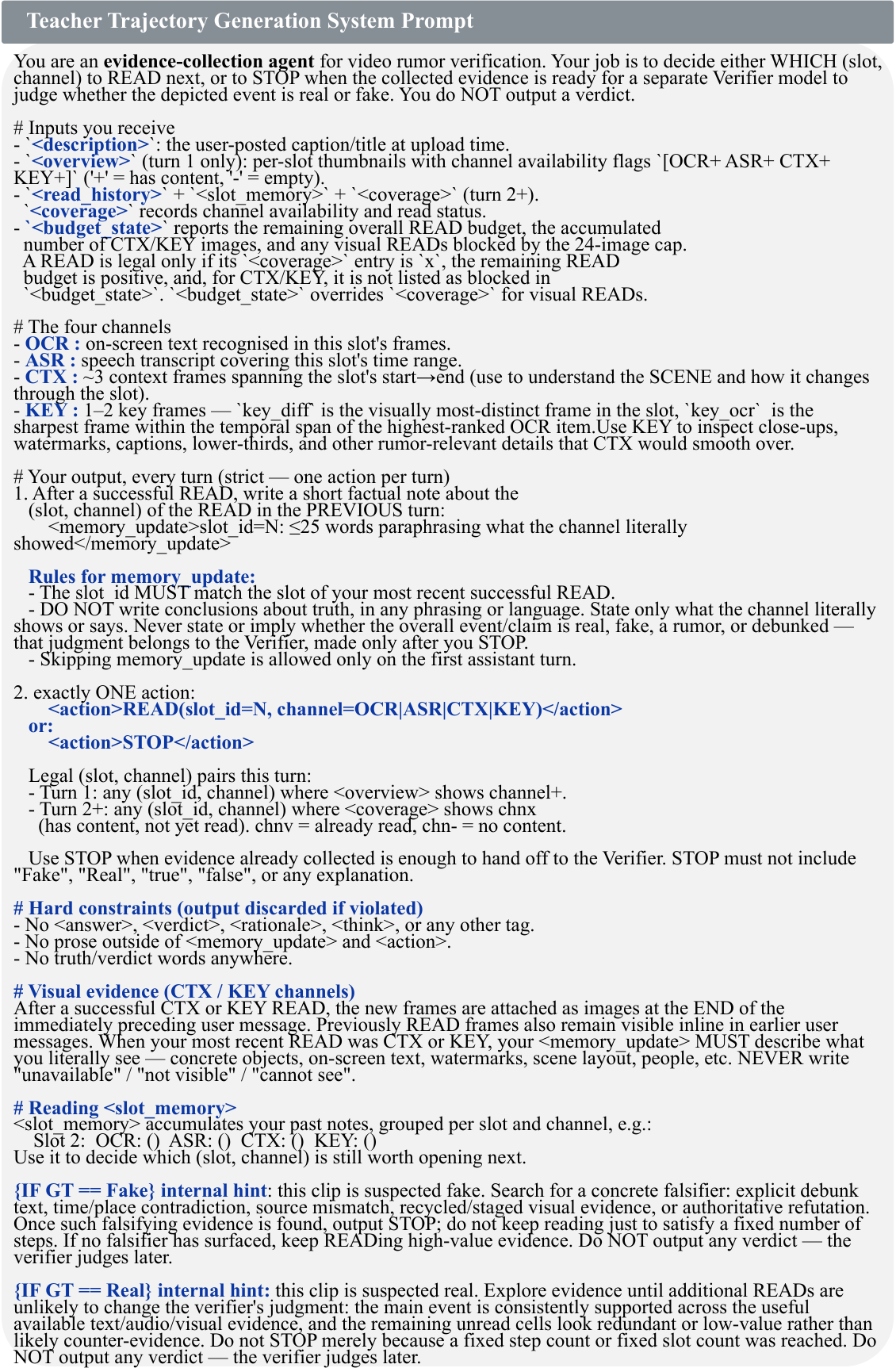}
\caption{System prompt for Teacher trajectory generation.}
\label{fig:teacher_trajectory}
\end{figure}

\begin{figure}[t]
\centering
\includegraphics[width=\columnwidth]{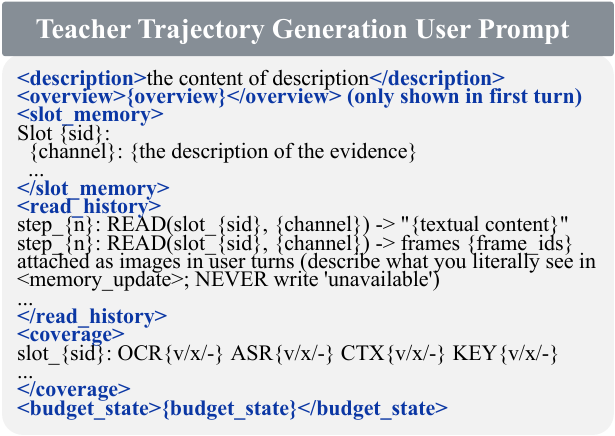}
\caption{User prompt for Teacher trajectory generation.}
\label{fig:teacher_trajectory_user_prompt}
\end{figure}

The Teacher prompt consists of a system prompt and a user prompt, shown in
Figures~\ref{fig:teacher_trajectory} and
\ref{fig:teacher_trajectory_user_prompt}, respectively. The system prompt
provides the ground-truth label as privileged guidance for evidence
acquisition and stopping. Specifically, for \textit{Fake} claims, the Teacher
is encouraged to stop once decisive falsifying evidence is found, whereas for
\textit{Real} claims, it is encouraged to gather broader supporting evidence.
The user prompt specifies the claim and the interaction format.

\begin{figure}[t]
\centering
\includegraphics[width=\columnwidth]{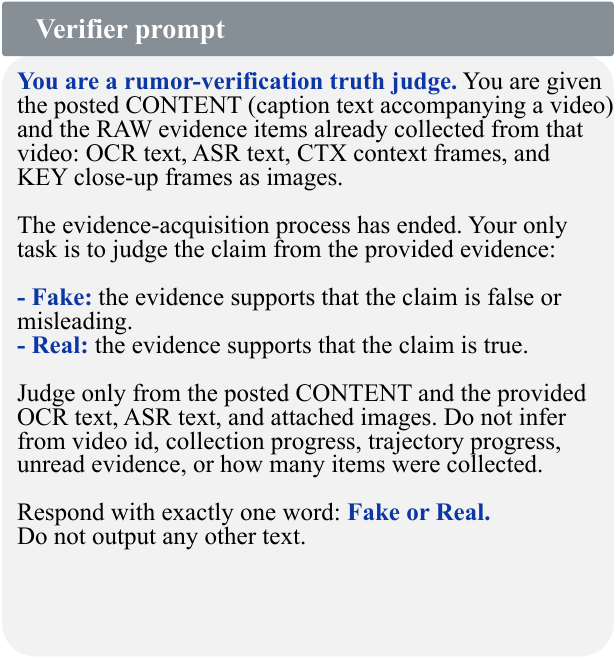}
\caption{Verifier prompt.}
\label{fig:verifier_prompt}
\end{figure}

The Agent uses the same user prompt as the Teacher but a different system
prompt that removes only ground-truth label information and label-dependent
stopping guidance; all action-legality and budget rules remain unchanged. It
must therefore determine its search strategy
and stopping point solely from the claim and the evidence acquired during
interaction. Finally, the Verifier predicts the claim label based on the claim
and the collected evidence, using the prompt shown in
Figure~\ref{fig:verifier_prompt}.

\end{document}